\pdfoutput=1

\documentclass[11pt]{article}

\usepackage[]{emnlp2022}

\usepackage{times}
\usepackage{latexsym}

\usepackage{graphicx}
\usepackage{subcaption}
\usepackage{latexsym}
\usepackage{lscape}
\usepackage{amsmath}
\usepackage{booktabs}
\usepackage{tabularx}
\usepackage{enumitem}
\usepackage{amssymb,amsmath,amsthm,enumitem}

\usepackage{amssymb}

\usepackage[T1]{fontenc}

\usepackage[utf8]{inputenc}

\newcommand{\spartqa}{\textsc{SpartQA}}
\newcommand{\spartun}{\textsc{SpaRTUN}}
\newcommand{\msprl}{\textsc{mSpRL}}
\newcommand{\sprl}{\textsc{SpRL}}
\newcommand{\resq}{\textsc{ReSQ}}
\newcommand{\auto}{\textsc{SpartQA-Auto}}
\newcommand{\human}{\textsc{SpartQA-Human}}
\newcommand{\stepgame}{\text{StepGame}}
\newcommand{\babi}{\text{bAbI}}
\newcommand{\bert}{\text{BERT}}

\usepackage[]{xcolor}
\usepackage{etoolbox}
\usepackage{multirow}
\usepackage{paralist}
\usepackage{soul}
\usepackage{hyperref}
\usepackage{amsmath}
\usepackage{latexsym}

\usepackage{microtype}

\usepackage{inconsolata}

\title{Transfer Learning with Synthetic Corpora for Spatial Role Labeling and Reasoning} 

\author{Roshanak Mirzaee \\
  Michigan State University \\
  \texttt{mirzaee@msu.edu} \\\And
  Parisa Kordjamshidi \\
  Michigan State University \\
  \texttt{kordjams@msu.edu} \\}

\begin{document}
\maketitle
\begin{abstract}

Recent research shows synthetic data as a source of supervision helps pretrained language models (PLM) transfer learning to new target tasks/domains. However, this idea is less explored for spatial language. 
We provide two new data resources on multiple spatial language processing tasks.
The first dataset is synthesized for transfer learning on spatial question answering (SQA) and spatial role labeling (SpRL). Compared to previous SQA datasets, we include a larger variety of spatial relation types and spatial expressions. Our data generation process is easily extendable with new spatial expression lexicons. The second one is a real-world SQA dataset with human-generated questions built on an existing corpus with \sprl{} annotations. This dataset can be used to evaluate spatial language processing models in realistic situations. 
We show pretraining with automatically generated data significantly improves the SOTA results on several SQA and \sprl{} benchmarks, particularly when the training data in the target domain  is small.

\end{abstract}

\section{Introduction}

Understanding spatial language is important in many applications such as navigation~\cite{zhang-kordjamshidi-2022-explicit, zhang-etal-2021-towards, chen2019touchdown}, medical domain
~\cite{datta2020understanding, kamel2019overview, massa2015machine}, and robotics 
~\cite{venkatesh2021spatial, kennedy2007spatial}. 
However, few benchmarks have directly focused on comprehending the spatial semantics of the text. Moreover, the existing datasets are either synthetic~\cite{mirzaee-etal-2021-spartqa, weston2015towards, shi2022stepgame}  or at small scale~\cite{mirzaee-etal-2021-spartqa,kordjamshidi2017clef}.  

\begin{figure}[t]
    \centering
    \includegraphics[width =\columnwidth]{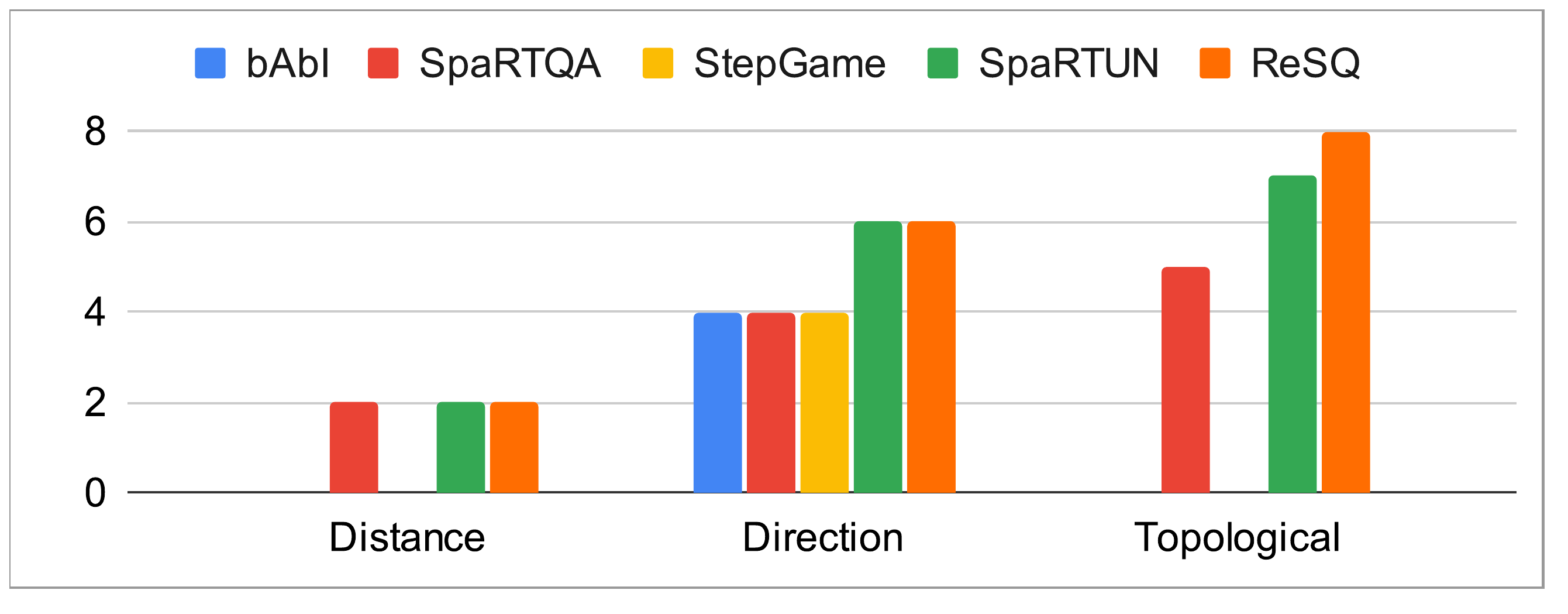}
    \caption{The comparative coverage of relation types based on Table~\ref{tab:spatial_relations} for SQA datasets.}
    \label{fig:coverage}
\end{figure}

The \textit{synthetic datasets} often focus on specific types of relations with a small coverage of spatial semantics needed for spatial language understanding in various domains. 
Figure~\ref{fig:coverage} indicates the coverage of sixteen spatial relation types~(in Table~\ref{tab:spatial_relations}) collected from existing resources~\cite{randell1992spatial, wolter2009sparq, renz2007qualitative}.
The \textit{human-generated datasets}, despite helping study the problem as evaluation benchmarks, are less helpful for training models that can reliably understand spatial language due to their small size~\cite{mirzaee-etal-2021-spartqa}. 
\begin{figure}[t]
	\centering
	\begin{subfigure}[t]{\linewidth}
	
		\includegraphics[width=\linewidth]{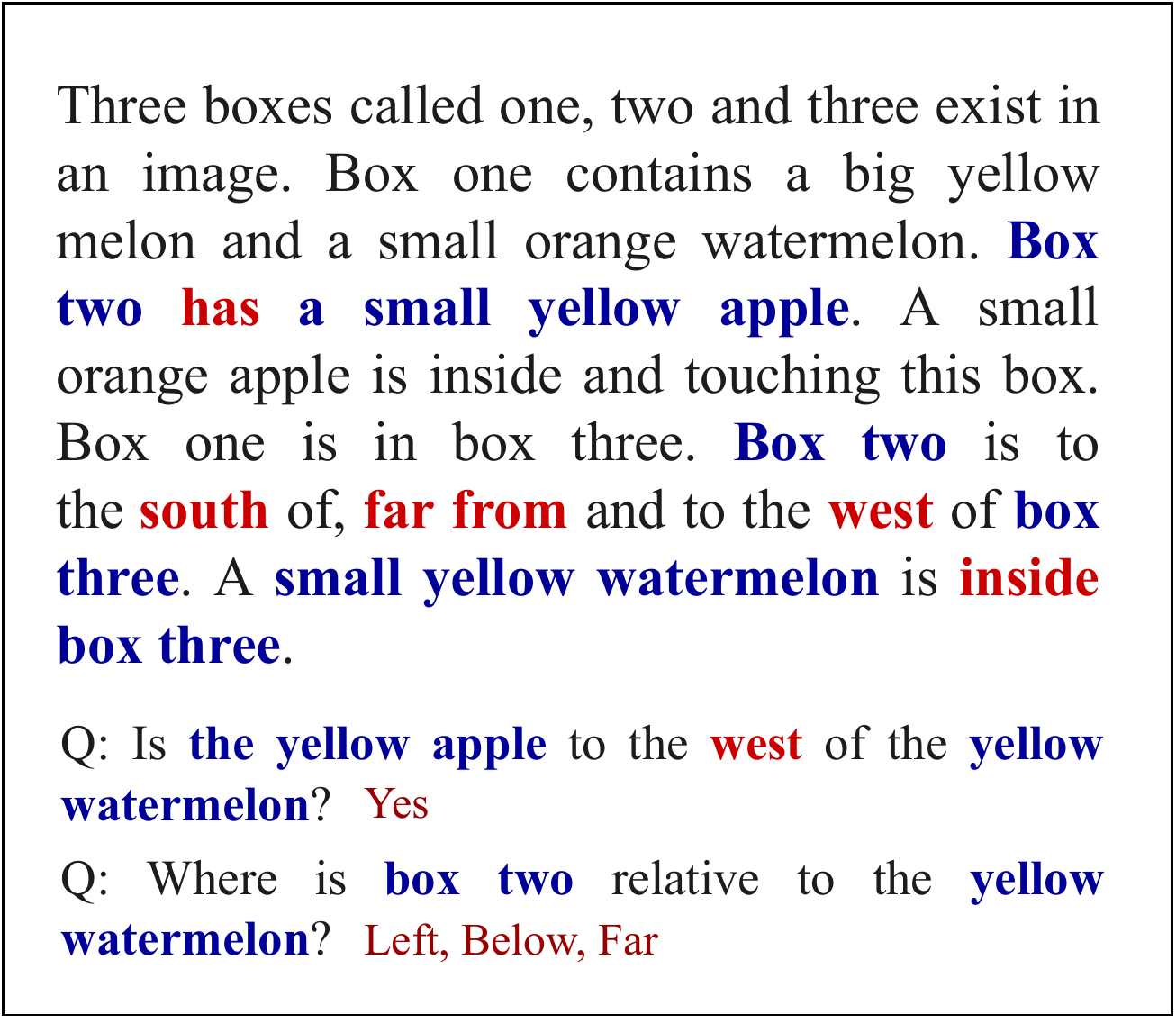}
		\caption{ \spartun{} - A synthetic large dataset
  provided as a source of supervision.
		}
		\label{fig:spartun}
	\end{subfigure}
	\begin{subfigure}[t]{\linewidth}
		\includegraphics[width=\linewidth]{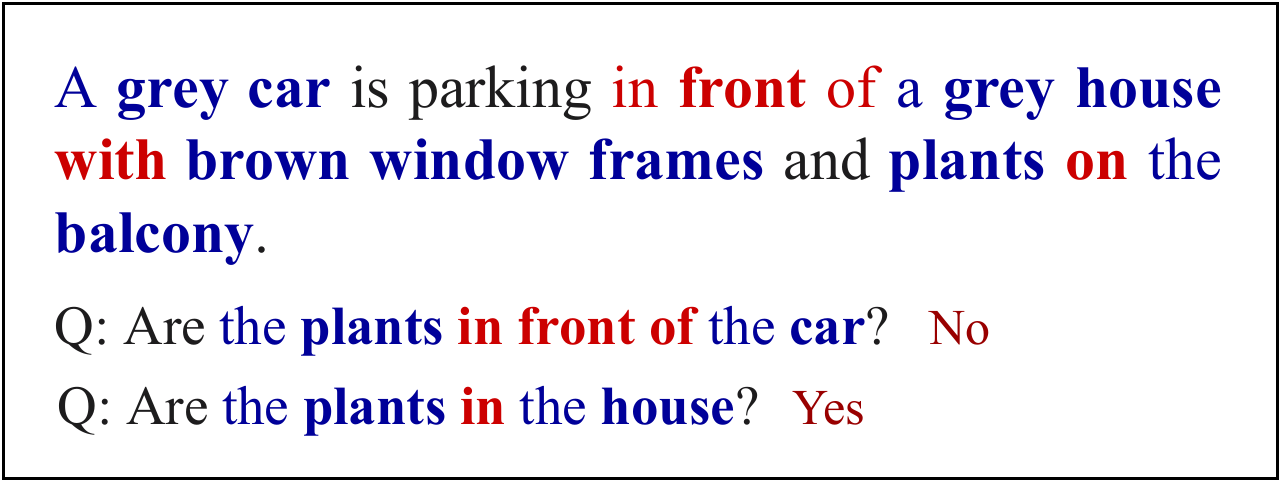}
		\caption{\resq{} - A human-generated dataset for probing the models on realistic SQA}
		\label{fig:sprlqa}
    \end{subfigure}
	\hfill
 	\caption{Two new datasets on SQA 
  }
	\label{fig:sqa}
\end{figure}
\begin{table*}[t]
\resizebox{\textwidth}{!}{%
\begin{tabular}{|l|l|l|l|}
\hline
\textbf{\begin{tabular}[c]{@{}l@{}}Formalism\\ (General Type)\end{tabular}} & \textbf{Specific value} & \textbf{Spatial type/Spatial value)} & \textbf{Expressions (e.g.)} \\ \hline
Topological & RCC8 & \begin{tabular}[c]{@{}l@{}}DC (disconnected)\\ EC (Externally Connected)\\ PO (Partially Overlapped)\\ EQ (Equal)\\ TPP (Tangential Proper Part)\\ NTPP (Non-Tangential Proper Part)\\ TPPI (Tangential Proper Part inverse)\\ NTPPI (Non-Tangential Proper Part inverse)\end{tabular} & \begin{tabular}[c]{@{}l@{}}disjoint\\ touching\\ overlapped\\ equal\\ covered by\\ in, inside\\ covers\\ has\end{tabular} \\ \hline
Directional & Relative & \begin{tabular}[c]{@{}l@{}}LEFT, RIGHT\\ BELOW, ABOVE\\ BEHIND, FRONT\end{tabular} & \begin{tabular}[c]{@{}l@{}}left of, right of\\ under, over\\ behind, in front\end{tabular} \\ \hline
Distance & Qualitative & Far, Near & far, close \\ 
\hline
\end{tabular}%
}
\caption{Spatial relation types and examples of spatial language expressions.}
\label{tab:spatial_relations}
\end{table*}

In this work, we build a new synthetic dataset on SQA, called \spartun{}\footnote{\textbf{Spa}tial \textbf{R}easoning and role labeling for \textbf{T}ext \textbf{UN}nderstanding}~(Fig~\ref{fig:spartun}) to provide a source of supervision with broad coverage of spatial relation types and expressions\footnote{We only consider explicit spatial semantics and the Metaphoric usages and implicit meaning are not covered in this work.}. 



To generate \spartun{}, we follow the idea of \spartqa{}~\cite{mirzaee-etal-2021-spartqa} benchmark and generate scene graphs from a set of images. 
The edges in this graph yield a set of triplets such as $\operatorname{ABOVE(blue\ circle, red\ triangle)}$,
which are used to generate a scene description (i.e., a story).

In \spartun{}, we map the spatial relation types in triplets (e.g., $\operatorname{ABOVE}$) to a variety of spatial language expressions~(e.g., over, north, above) to enable the transfer learning for various data domains
\footnote{The full list of spatial expressions used in this dataset and the dataset generation code are provided in \url{https://github.com/HLR/SpaRTUN}.}.
We also build a logical spatial reasoner to compute all possible direct and indirect spatial relations between graph nodes. Then, the questions of this dataset are selected from the indirect relations.

To evaluate the effectiveness of \spartun{} in transfer learning, we created another dataset named \resq{}\footnote{\textbf{Re}al-world \textbf{S}patial \textbf{Q}uestions}~(Fig~\ref{fig:sprlqa}).
This dataset is built on \msprl{}~\cite{kordjamshidi2017clef} corpus while we added human-generated spatial questions and answers to its real image descriptions. This dataset comparatively reflects more realistic challenges and complexities of the SQA problem.

We analyze the impact of \spartun{} as source of extra supervision on several SQA and \sprl{} benchmarks.
To the best of our knowledge, we are the first  to use synthetic supervision for the \sprl{} task. 
Our results show that the auto-generated data successfully improves the SOTA results on \msprl{} and \human{}, which are annotated for \sprl{} task.
Moreover, further pretraining models with \spartun{} for SQA task improves the result of previous models on \resq{}, \stepgame{}, and \human{} benchmarks. 
Furthermore, studying the broad coverage of spatial relation expressions of \spartun{} in realistic domains demonstrates that this feature is a key factor for transfer learning.


The contributions of this paper can be summarized as:
\textbf{(1)} We build a new synthetic dataset to serve as a source of supervision and transfer learning for spatial language understanding tasks with broad coverage of spatial relation types and expressions~(which is easily extendable); \textbf{(2)} We provide a human-generated dataset to evaluate the performance of transfer learning on real-world spatial question answering; \textbf{(3)} We evaluate the transferability of the models pretrained with \spartun{} on multiple SQA and \sprl{} benchmarks and show significant improvements in SOTA results.

\section{Related Research}


Requiring large amounts of annotated data is a well-known issue in training complex deep neural models~\cite{zhu2016we}
that is extended to spatial language processing tasks.
In our study, we noticed that all available large datasets on SQA task including \babi{}~\cite{weston2015towards}, \auto{}~\cite{mirzaee-etal-2021-spartqa}, and \stepgame{}~\cite{shi2022stepgame} are, all, synthetic.

\babi{} is a simple dataset that covers a limited set of relation types, spatial rules, and vocabulary. \stepgame{} focuses on a few relation types but with more relation expressions for each and considers multiple reasoning steps. \auto{}, comparatively, contains more relation types and needs complex multi-hop spatial reasoning. However, it contains a single linguistic spatial expression for each relation type.
All of these datasets are created based on controlled toy settings and are not comparable with real-world spatial problems in the sense of realistic language complexity and coverage of all possible relation types. 
\human{}~\cite{mirzaee-etal-2021-spartqa} is a human-generated version of \auto{} with more spatial expressions.  However, this dataset is provided for probing purposes and has a small training set that is not sufficient for effectively training deep models.

For the \sprl{} task, \msprl{} and SpaceEval~(SemEval-2015 task 8)~\cite{pustejovsky2015semeval} are two available datasets with spatial roles and relation annotations. These are small-scale datasets for studying the \sprl{} problem. 
From the previous works which tried transfer learning on \sprl{} task,
\cite{moussa2021spatial} only used it on word embedding of their \sprl{} model, and \cite{shin2020bert} used PLM without any specifically designed dataset for further pretraining.
These issues motivated us to create \spartun{} for further pretraining and transfer learning for SQA and \sprl{}. 

Transfer learning has been used effectively in different NLP tasks to further fine-tune the PLMs~\cite{razeghi2022impact, alrashidi2020automatic, magge2018deep}.
Besides transfer learning, several other approaches are used to tackle the lack of training data in various NLP areas, such as providing techniques to label the unlabeled data~\cite{enayati-etal-2021-visualization}, using semi-supervised models~\cite{van2019semi, li2021semi} or data augmentation with synthetic data~\cite{li2019logic, min2020syntactic}.
However, transfer learning is a simple way of using synthetic data as an extra source of supervision at no annotation cost. Compared to the augmentation methods, the data in the transfer learning only needs to be close to the target task/domain~\cite{ma2021knowledge} and not necessarily the same. 
\citeauthor{mirzaee-etal-2021-spartqa} is the first work that considers transfer learning for SQA. It shows that training models on synthetic data and finetuning  with small human-generated data results in a better performance of PLMs. However, their coverage of spatial relations and expressions is insufficient for effective transfer learning to realistic domains.

 
Using logical reasoning for building datasets that need complex reasoning for question answering is used before in building QA datasets~\cite{ijcai2020p0537, saeed2021rulebert}. More recent efforts even use the path of reasoning and train models to follow that~\cite{tafjord2020proofwriter}. However, there are no previous works to model spatial reasoning as we do here with the broad coverage of spatial logic.


\section{Transfer Learning for Spatial Language Understanding}

\label{sec:tasks}

To evaluate transfer learning on spatial language understanding, we select two main tasks, spatial question answering (SQA) and spatial role labeling (\sprl{}). 
Given the popularity of PLMs in transfer learning~\cite{khashabi2020unifiedqa, ma2021knowledge, ijcai2020p0537},  we design PLM-based models for this evaluation. In the rest of this section, we describe each task and model in detail.

\subsection{Spatial Question Answering}

In spatial question answering, given a scene description, the task is to answer questions about 
the spatial relations between entities~(e.g., Figure~\ref{fig:sqa}).
Here, we focus on challenging questions that need multi-hop spatial reasoning over explicit relations. We consider two question types, YN~(Yes/No) and FR(Find relations). The answer to YN  is chosen from "Yes" or "No," and the answer to FR is chosen from a set of  relation types.




We use a PLM with classification layers as a baseline for the SQA task. 
We use a binary
classification layer for each label for questions with more than one valid answer and a multi-class classification layer for questions with a single valid answer.  
To predict the answer, we pass the concatenation of the question and story to the PLM~(more detail in \cite{devlin-etal-2019-bert}.)
The final output of $[CLS]$ token is passed to the classification layer and depending on the question type, a label or multiple labels with the highest probability are chosen as the final answer.

We train the models based on the summation of the cross-entropy losses of all binary classifiers in multi-label classification  or the single cross-entropy for a single classifier in multi-classification. In the multi-label setting, we remove inconsistent answers by post-processing during the inference phase. For instance, LEFT and RIGHT relations cannot be valid answers simultaneously.

\subsection{Spatial Role Labeling}

\begin{figure}[t]
    \centering
    \includegraphics[width=\linewidth]{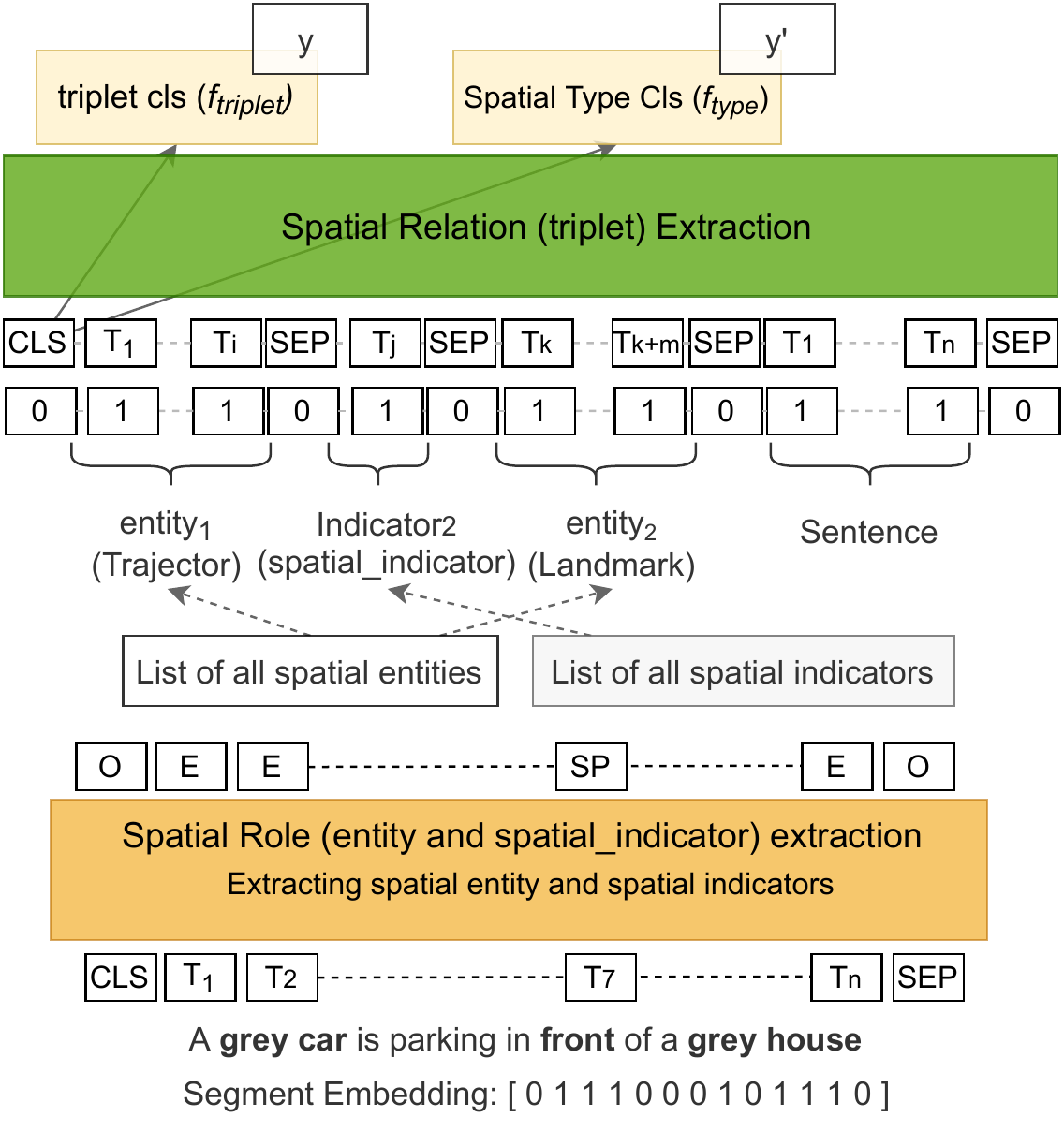}
    \caption{Spatial role labeling model includes two separately trained modules. E: entity, SP: spatial\_indicators. As an example, triplet (a grey house,  front , A grey car) is correct and the ``spatial\_type = FRONT'', and (A grey car,  front, a grey house) is incorrect, and the ``spatial\_type = NaN''.
    }
    \label{fig:pipeline}
\end{figure}

 Spatial role labeling~\cite{kordjamshidi2010spatial,kordjamshidi2011spatial} is
 the task of identifying and classifying the spatial roles~(Trajector, Landmark, and spatial indicator) and their relations.  A relation is selected from the relation types in Table \ref{tab:spatial_relations} and assigned to each triplet of $(\operatorname{Trajector},\ \operatorname{Spatial\ indicator},\ \operatorname{Landmark})$ extracted from the sentence. We call the former \textbf{spatial role extraction} and the latter \textbf{spatial relation\footnote{In different works like \cite{kordjamshidi2010spatial}, the triplet and relation are used interchangeably.} extraction}~(Figure~\ref{fig:pipeline}).
 
 
Several neural models have been proposed to solve spatial role~\cite{mazalov2015spatial, ludwig2016deep, datta2020hybrid}.
We take a similar approach to prior research \cite{shin2020bert} for the extraction of spatial roles~(entities~(Trajector/Landmark) and spatial indicators).

First, we separately tokenize each sentence in the context and use a PLM~(which is \bert{} here) to compute the tokens representation. 
Next, we apply a BIO tagging layer on tokens representations
using (O, B-entity, I-entity, B-indicator, I-indicator)  tags. A Softmax layer on BIO tagger output is used to select the spatial entities and spatial indicators with the highest probability. For training, we use CrossEntropy loss given the spatial annotation.

For the spatial relation extraction model, similar to~\cite{yao2019kg,shin2020bert},  we use \bert{} and a classification layer to extract correct triplets.
Given the output of the spatial role extraction model, for each combination of $(\operatorname{spatial\ entity} (tr),\ \operatorname{spatial\_indicator} (sp),$ $\ \operatorname{spatial\ entity} (lm))$ in each sentence we create an input\footnote{$[\text{CLS}, tr, \text{SEP}, sp, \text{SEP}, lm, \text{SEP}, sentence, \text{SEP}]$ } and pass it to the \bert{} model. To indicate the position of each spatial role in the sentence, we use segment embeddings and add $1$ if it is a role position and $0$ otherwise.

The $[CLS]$ output of \bert{} will be passed to a one-layer MLP that provides the probability for the triplet~(see Fig~\ref{fig:pipeline}).
Compared to the prior research, we  predict the spatial type for each triplet as an auxiliary task for spatial relation extraction. To this aim, we apply another multi-class classification layer\footnote{The classes are relation types in Table~\ref{tab:spatial_relations} alongside a $\operatorname{NaN}$ class for incorrect triplets.} on the same $[CLS]$ token. To train the model, we use a joint loss function for both relation and type 
modules~(more detail in Appendix~\ref{sec:setup}).


\section{\spartun{}: Dataset Construction}
\label{sec:spartun}

\begin{figure*}[th]
    \centering
    \includegraphics[width = \textwidth]{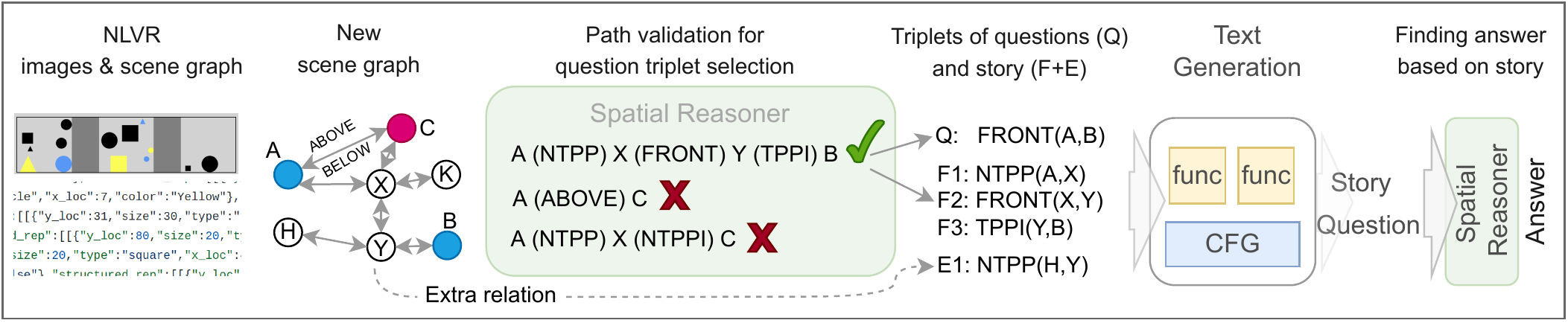}
    \caption{The data construction flow of \spartun{}. First, we generate scene graphs from NLVR images. Then a spatial reasoner validates each path between each pair of entities in this graph. All facts~($F$) in the selected $path$ and some extra facts~($E$) from the scene graph are selected as story triplets, and the start and end nodes of the $path$ are selected as question triplets. Finally, we pass all triplets to a text generation module and compute the final answer. We ignore paths with length one~(e.g., $A(\operatorname{}{ABOVE})C$) and only keep questions that need multi-hop reasoning.}
    \label{fig:spartun-generation}
\end{figure*}

To provide a source of supervision for spatial language understanding tasks, we generate a synthetic dataset with SQA format that contains \sprl{} annotation of sentences. We build this dataset by expanding \spartqa{} in multiple aspects. The following additional features are considered in creating \spartun{}: 

\noindent F1)  A broad coverage of  various types of spatial relations and
including rules of reasoning over their combinations (e.g. 
$\operatorname{NTPP}(a, b),\ \operatorname{LEFT}(b,c) \rightarrow{} \operatorname{LEFT}(a,c)$
) in various domains. 

\noindent F2) A broad coverage of spatial language expressions and utterances used in various domains.

\noindent F3) Including extra annotations such as the supporting facts and number of reasoning steps for SQA to be used in complex modeling.

In the rest of this section, we describe the details of creating \spartun{}  and the way we support the above mentioned features. Figure~\ref{fig:spartun-generation} depicts \spartun{} data construction flow.


\paragraph{Spatial Relation Computation.}

Following \auto{}, we use the NLVR scene graphs~\cite{suhr2017corpus}
and compute relations between objects in each block based on their given coordinates. NLVR is limited to 2D relation types\footnote{The relations types included in NLVR are: DC, EC, *PP relations, LEFT, RIGHT, BELOW, and ABOVE.}, therefore to add more dimensions (FRONT and BEHIND), we randomly change the LEFT and RIGHT to BEHIND and FRONT in a subset of examples. Moreover, there are no relations between blocks in NLVR descriptions. 

To expand the types of relations, we extend this limitation and randomly assign relations\footnote{All relation in Table~\ref{tab:spatial_relations} except EQ} to the blocks while ensuring the spatial constraints are not violated. 
Then, we create a new scene graph with computed spatial relations. The nodes in this graph represent the entities (objects or blocks), and the directed edges are the spatial relations. 

\paragraph{Question Selection.} 
There are several paths between each pair of entities in the generated scene graph. We call a path valid if at least one relation can be inferred between its start and end nodes can be inferred.
For example, in Figure~\ref{fig:spartun-generation}, $\operatorname{NTPP}(A,X), \ \operatorname{FRONT}(X,Y), \ \operatorname{TPPI}(Y,B)$ is valid since it results in $\operatorname{FRONT}(A, B)$ while $\operatorname{NTPP}(A,X),\ \operatorname{NTPPI}(X,C)$ is not a valid path --there is no rules of reasoning that can be applied to infer new relations.

To  verify the validity of each path, we pass its edges, represented as triplets in the predicate-arguments form to a logical spatial reasoner~(implemented in Prolog) and query all possible relations between the pair. The number of triplets in each path represents the number of reasoning steps for inferring the relation.  

We generate the question triplets from the paths with the most steps of reasoning~(edges). This question will ask about the spatial relationship between the head and tail entity of the selected path.
The triplets in this path are used to generate the story
and are annotated as supporting facts. Additionally, the story will include additional information~(extra triplets) unnecessary for answering the question to increase the complexity of the task. 

\begin{table*}[th]
\resizebox{\textwidth}{!}{%

\begin{tabular}{|llllll|}
\hline
&&&&& \\
Not          & $\forall (X,Y) \in Entities$     & $R \in \left \{ Dir \vee PP \right \}$         &  & $\operatorname{IF} R(X,Y)$                       & $\Rightarrow \operatorname{NOT}(R\_reverse(X,Y))$                  \\
Inverse      & $\forall (X,Y) \in Entities$     & $R \in \left \{ Dir \vee PP \right \}$ &  & $\operatorname{IF} R(Y,X)$                       & $\Rightarrow R\_reverse(X,Y)$                                      \\
Symmetry     & $\forall (X,Y) \in Entities$     & $R \in \left \{ Dis \vee (RCC - PP) \right \}$ &  & $\operatorname{IF} R(Y,X)$                       & $\Rightarrow R(X,Y)$                                               \\
Transitivity & $\forall (X,Y, Z) \in Entities$  & $R \in \left \{ Dir \vee PP \right \}$         &  & $\operatorname{IF} R(X,Z), R(Z,Y) $              & \begin{tabular}[c]{@{}l@{}}$\Rightarrow  R(X,Y)$\end{tabular} \\
Combination  & $\forall (X,Y,Z,H) \in Entities$ & $R \in Dir, *PP \in PP $                       &  & $\operatorname{IF} *PP(X,Z), R(Z,H), *PPi(Z,Y) $ & \begin{tabular}[c]{@{}l@{}}$\Rightarrow  R(X,Y)$ 
\end{tabular}\\
&&&&&\\
\hline
\end{tabular}%
}
\caption{Designed spatial rules.  $Dir$: Directional relations~(e.g., LEFT), $Dis$: Distance relations~(e.g., FAR), $PP$: all Proper parts relations~(NTPP, NTPPI, TPPI, TPP), $RCC-PP$: All RCC8 relation except proper parts relations. $*PP$: one of TPP or NTPP. $*PPi$: one of NTPPi or TPPi.}
\label{fig:sp-rules}
\end{table*}

\paragraph{Spatial Reasoner.}
We implement several rules~(in the form of Horn clauses shown in Table~\ref{fig:sp-rules}) in Prolog, which express the logic between the relation types~(described in Table~\ref{tab:spatial_relations}) in various formalisms and model the logical spatial reasoning computation~(see Appendix~\ref{sec: prolog}). Compared to previous tools~\cite{wolter2009sparq}, we are the first to include the spatial, logical computation between multiple formalisms.  This reasoner validates the question/queries based on the given facts. 
For instance, by using the Combination rule
in Table~\ref{fig:sp-rules} over the set of facts $\{\operatorname{NTPP}(A,X),\ \operatorname{FRONT}(X,Y),\ \operatorname{TPPI}(Y,B)\}$, the reasoner returns \textit{True} for the query $\operatorname{FRONT}(A,B)$ and \textit{False} for $\operatorname{FRONT}(B,A)$ or $\operatorname{BEHIND}(A,B)$.

\paragraph{Text generation.} 
The scene description is generated from the selected story triplets in question selection phase and using a publicly available context-free grammar~(CFG) provided in \auto{}.
However, we increase the variety of spatial expressions 
by using a vocabulary of various entity properties and relation expressions~(e.g., above, over, or north for ABOVE relation type) taken from existing resources~\cite{freeman1975modelling, mark1989languages, lockwood2006automatic, stock2022detecting, Herskovits86}
We map the relation types and the entity properties to the lexical forms in our collected vocabulary.

For the question text, we generate the entity description and relation expression for each question triplet.
The entity description is generated based on a subset of its properties in the story. 
For instance, an expression such as ``a black object'' can be generated to refer to both ``a big black circle'' and ``a black rectangle''.
We generate two question types, YN~(Yes/No) questions that ask whether a specific relation exists between two entities, and FR~(Find Relations) questions that ask about all possible relations between them. To make YN questions more complex, we add quantifiers (``all'' and ``any'') to the entities' descriptions.

Our text generation method can flexibly use an extended vocabulary to provide a richer corpus to supervise new target tasks when required. 

\paragraph{Finding Answers.}
We search all entities in the story based on the entity descriptions~(e.g., all circles, a black object) in each question and use the spatial reasoner to find the final answer.


\paragraph{SpRL Annotations.}
 Along with generating the sentences for the story and questions, we automatically annotate the described spatial configurations with spatial roles and relations (trajector, landmark, spatial indicator, spatial type, triplet, entity ids). These annotations are based on a previously proposed annotation scheme of \sprl{} and provide free annotations for the \sprl{} task.


To generate \spartun{}, we use 6.6k NLVR scene graphs for  training and 1k for each dev and test set. We collect 20k training, 3k dev, and 3k test examples for each FR and YN question~(see Table~\ref{tab:statistical})\footnote{All data are provided in the English language.: The corpus is in English.}.
On average, each story of \spartun{} contains eight sentences and 91 tokens that describe on average 10 relations between different mentions of entities. 
More details about the dataset statistics can be seen in Appendix~\ref{sec:appendix-spartun}. 



\section{Experimental Results}
\label{sec:experiments}
The focus of this paper is to provide a generic source of supervision for spatial language understanding tasks rather than proposing new techniques or architectures. Therefore, in the experiments, we analyze the impact of \spartun{} on SQA and \sprl{} using the PLM-based models described in Section~\ref{sec:tasks}.

In all experiments, we compare the performance of models \textit{fine-tuned with the target datasets} with and without \textit{further pretraining on \textbf{syn}thetic \textbf{sup}ervision}~(\textit{SynSup}). All codes are publicly available\footnote{\url{https://github.com/HLR/Spatial-QA-tasks}}. The details of experimental settings and  hyperparameters of datasets are provided in the Appendix.

\subsection{Spatial Question Answering}

Here, we evaluate the impact of \spartun{} and compare it with the supervision received from other existing synthetic datasets.
Since the datasets that we use contain different question types, we supervise the models based on 
the same question type as the target task\footnote{\stepgame{} only has FR question types. Hence, we use the model trained on FR questions for both FR and YN target tasks.}.

The baselines for all experiments include a majority baseline~(MB) which predicts the most repeated label as the answer to all questions, and a pretrained language model, that is, \bert{} here. We also report the human accuracy in answering the questions for the human-generated datasets\footnote{All human results gathered by scoring the human answers over a subset of the test set.}. 
For all experiments, to evaluate the models, we measure the accuracy which is the percentage of correct predictions in the test sets.

\subsubsection{SQA Evaluation Datasets}

\begin{table}[h]

\centering
\resizebox{\columnwidth}{!}{\begin{tabular}{|l|ccc|}
\hline
\textbf{Dataset} & \multicolumn{1}{l}{\textbf{Train}} & \multicolumn{1}{l}{\textbf{Dev}} & \multicolumn{1}{l|}{\textbf{Test}} \\ \hline
\babi{} & 8992 & 992 & 992  \\ \cline{1-1}
\auto{}~(YN) & 26152 &3860  & 3896   \\
\auto{}~(FR) & 25744 &3780  & 3797   \\ \cline{1-1}
\multicolumn{1}{|l|}{\human{}~(YN)}& 162 & 51 & 143 \\
\multicolumn{1}{|l|}{\human{}~(FR)}& 149 & 28 & 77  \\ \cline{1-1}
\resq{} & 1008 & 333 & 610 \\\cline{1-1}
\stepgame{} & 50000& 1000 & 10000\\\cline{1-1}
\spartun{}~(YN) & 20334 &3152& 3193\\
\spartun{}~(FR) & 18400 &2818& 2830\\

\hline
\end{tabular}}
\caption{Size of SQA benchmarks.}
\label{tab:statistical}
\end{table}

\paragraph{\babi{}} We use tasks 17 and 19 of \babi{}. Task 17 is on spatial reasoning and contains binary Yes/No questions. Task 19 is on path finding and contains FR questions with answers in \{LEFT, RIGHT, ABOVE, BELOW\} set. The original dataset contains west, east, north, and south, which we mapped to their corresponding relative relation type.

\paragraph{\human} is a small human-generated dataset containing YN and FR questions that need multi-hop spatial reasoning. 
The answer of YN questions is in \{Yes, No,DK\} where DK denotes \textit{Do not Know} is used when the answer cannot be inferred from the context. 
The answer to FR questions is in  \{left, right, above, below, near to, far from, touching, DK\}\footnote{Since the relation types are not used in \spartqa{}, the answer is selected from a fixed set of relation expressions}.

\paragraph{\stepgame{}} is a synthetic SQA dataset containing FR questions which need $k$ reasoning steps to be answered ($k = 1$ to $10$).
The answer to each question is one relation in \{left, right, below, above, lower-left, upper-right, lower-right, upper-left\} set. 

\paragraph{ \resq{}} We created this dataset to reflect the natural complexity of real-world spatial descriptions and questions. We asked three volunteers (English-speaking undergrad students) to generate Yes/No questions for \msprl{} dataset
that contains complex human-generated sentences. 
The questions require at least one step of reasoning. The advantage of \resq{} is that the human-generated spatial descriptions and their spatial annotations already exist in the original dataset. The statistics of this dataset are provided in Appendix~\ref{sec:sprlqa}. 

One of the challenges of the \resq{}, which is not addressed here,  is that the questions require spatial commonsense knowledge in addition to capturing the spatial semantics. For example, by using commonsense knowledge from the sentence, ``a lamp hanging on the ceiling'', we can infer that the lamp is above all the objects in the room. 
To compute the human accuracy, we asked two volunteers to answer 100 questions from the test set of \resq{} and compute the accuracy. 



\subsubsection{Transfer Learning in SQA}
The following experiments demonstrate the impact of transfer learning for SQA benchmarks considering different supervisions.

\begin{table}[]
\tiny
\centering
\resizebox{\columnwidth}{!}{\begin{tabular}{|llcc|}
\hline
 \textbf{Model}& \textit{\textbf{SynSup}} & \textbf{17$^{\operatorname{1k}}$}& \textbf{19$^{\operatorname{500}}$} \\ \hline
MB & - & 51.9& 10.6 \\ \hline
BERT & - & 87.39& 34.53  \\ \hline
BERT &\spartqa{}-A & 90.42 & \textbf{100} \\
BERT &\stepgame{} & 87.39  & 99.89 \\
BERT & \spartun{}-S & \textbf{92.43} & 98.99 \\
BERT & \spartun{} & 90.02 &  99.89\\ \hline
\end{tabular}}
\caption{Impact of using synthetic supervision on the \babi{} tasks. All the models are further fine-tuned on the training set of task 17~(size = 1k) and 19~(size = 500), and test on \babi{} test sets.}
\label{tab:babi}
\end{table}

\begin{table}[]
\tiny
\centering
\resizebox{\columnwidth}{!}{\begin{tabular}{|llcc|}
\hline
 \textbf{Model}& \textit{\textbf{SynSup}} & \textbf{YN}& \textbf{FR} \\ \hline
MB & - & 53.60 & 24.52 \\ \hline
BERT & - &\textbf{49.65} & 18.18  \\ \hline
BERT &\spartqa{}-A&  39.86 &  48.05\\
BERT &\stepgame{}& 44.05  & 11.68 \\
BERT & \spartun{}-S & 44.75 &  37.66\\
BERT & \spartun{} & 48.25 &  \textbf{50.64}\\ \hline
Human &  - & 90.69 & 95.23 \\ \hline
\end{tabular}}
\caption{Transfer learning on \human{}. \spartqa{}-A stands for \auto{}.
}
\label{tab:human-qa}
\end{table}

\begin{table*}[t]
    
    \centering
    \resizebox{\textwidth}{!}{
    \begin{tabular}{|ll|cccccccccc|}
        \hline
        \multicolumn{1}{|c}{}& \multicolumn{1}{c|}{}& 
        \multicolumn{10}{c|}{k steps of reasoning} \\
        \hline
        Model&
        SynSup & 1 &2&3&4&5&6&7&8&9&10\\ \hline
        
         TP-MANN & - & 
         85.77&	60.31&	50.18&	37.45&	31.25&	28.53&	26.45&	23.67&	22.52&	21.46
        \\ \hline
         
         \bert{}&- & 98.44&	94.77&	91.78&	71.7&	57.56&	50.34&	45.17&	39.69&	35.41&	33.62 \\ \hline

         \bert{}&\spartqa{}-A & 
         98.63&	94.95&	91.94&	77.74&	68.37&	61.67&	57.95&	50.82&	46.86&	44.03
         \\ 
         \bert{}&\spartun{}-S &\textbf{98.70} & \textbf{95.21}  & \textbf{92.46} & 77.93 & 69.53   &  62.14 & 57.37 & 48.79&  44.67  & 42.72\\
         \bert{} & \spartun{} &  98.55   &  95.02   & 92.04  & \textbf{79.1 }  &      \textbf{70.34}& \textbf{ 63.39}   &   \textbf{58.74}  &    \textbf{52.09}  & \textbf{48.36}       & \textbf{45.68}\\ \hline
         
    \end{tabular}
    }
    \caption{Result of models with and without extra synthetic supervision on \stepgame{}.}
    \label{tab:stepgame}
\end{table*}



 Due to the simplicity of \textbf{bAbI} dataset, 
 PLM can solve this benchmark with 100\% accuracy~\cite{mirzaee-etal-2021-spartqa}. Hence we run our experiment on only  1k and 500 training examples of task 17 and task 19, respectively. 
Table~\ref{tab:babi} demonstrates the impact of synthetic supervision on both tasks of \babi{}. The results with various synthetic data are fairly similar for these two tasks. 
However, pretraining the model with the simple version of \spartun{}, named \spartun{}-S, performs better than other synthetic datasets on task 17. This can be due to the fewer relation expressions in \spartun{}-S, which follows the same structure as task 17. 



In the next experiment, we investigate the impact of \spartun{} on \textbf{\human{}} result. Comparing the results in Table~\ref{tab:human-qa}, we find that even though the classification layer for \auto{} and \human{} are the same, the model trained on \spartun{} has a better transferability. It achieves $2.6$\% better accuracy on FR and $9$\% better accuracy on YN questions compared to \auto{}.  YN is, yet, the most challenging question type in \human{} and none of the PLM-based models can reach even the simple majority baseline. 


Table~\ref{tab:stepgame} demonstrates our experiments on \textbf{\stepgame{}}. \bert{} without any extra supervision, significantly, outperforms the best reported model in ~\citeauthor{shi2022stepgame}, TP-MANN, which is based on  a neural memory network. As expected, all the PLM-based models almost solve the questions with one step of reasoning~(i.e. where the answer directly exists in the text). However, with increasing the steps of reasoning, the performance of the models decreases. Comparing the impact of different synthetic supervision, \spartun{} achieves the best result on $k>3$. For questions with $k<=3$, \spartun{}-S achieves competitive similar results compared to \spartun{}. Overall, the performance gap in \spartun{}-S, \auto{} and \spartun{} shows that more coverage of relation expressions in \spartun{} is effective.

\begin{table}[]
\small
\centering
\resizebox{0.85\columnwidth}{!}{\begin{tabular}{|llc|}
\hline
 \textbf{Model}& \textit{\textbf{SynSup}} & \textbf{Accu}\\ \hline
MB& - & 50.21  \\ \hline
BERT & - & 57.37 \\ \hline
BERT &\auto{}&  55.08   \\
BERT &\stepgame{}&  60.14   \\
BERT & \spartun{}-S  & 58.03 \\
BERT & \spartun{} & \textbf{63.60}  \\ \hline
Human & - & 90.38 \\ \hline
\end{tabular}}
\caption{Results with and without extra supervision on ReSQ. The Human accuracy is the performance of human on answering a subset of test set.
}
\label{tab:sprlqa}
\end{table}

In the next experiment, we show the influence of \spartun{} on real-world examples, which contain more types of spatial relations and need more rules of reasoning to be solved. Table~\ref{tab:sprlqa} shows the result of transfer learning on\textbf{ \resq{}}. 
This result shows that the limited coverage of spatial relations and expression in \auto{} impacts the performance of \bert{} negatively.
However, further pretraining \bert{} on \spartun{}-S improves the result on \resq{}. This can be due to the higher coverage of relation types in \spartun{}-S than \auto{}.
Using \spartun{} for further pretraining \bert{} has the best performance and improves the result by $5.5$\%, indicating its advantage for transferring knowledge to solve real-world spatial challenges.


\subsection{Spatial Role Labeling}


Here, we analyze the influence of the extra synthetic supervision on \sprl{} task when evaluated on human-generated datasets. Table~\ref{tab:sprlstatistical} shows the number of sentences in each \sprl{} benchmarks.

The pipeline model provided in Section~\ref{sec:tasks}, contains two main parts, a model for spatial role extraction~(SRol) 
and a model for spatial relation extraction~(SRel), which we analyze separately. 

We further pretrain the \bert{} module in these models and then fine-tune it on the target domain. We use Macro F1-score (mean of F1 for all classes) to evaluate the performance of the SRol and SRel models.

\subsubsection{\sprl{} Evaluation Datasets}
\begin{table}[th]

\centering
\resizebox{\columnwidth}{!}{\begin{tabular}{|l|ccc|}
\hline
\textbf{Dataset} & \multicolumn{1}{l}{\textbf{Train}} & \multicolumn{1}{l}{\textbf{Dev}} & \multicolumn{1}{l|}{\textbf{Test}} \\ \hline
\auto{}~(story) & 25755 &16214  & 16336   \\
\auto{}~(question) & 23584 &15092  & 15216   \\
\cline{1-1}
\human{}~(story)& 176 &99  & 272  \\
\human{}~(question) & 155 &127  & 367   \\
\cline{1-1}
\spartun{}~(story) & 48368 &7031& 7191\\
\spartun{}~(question) & 38734 &5970& 6023\\
\cline{1-1}
\msprl{}& 481 & -  & 461  \\ 
\hline
\end{tabular}}
\caption{Number of sentences of \sprl{} benchmarks. To train the \auto{}, we only use the 3k training examples~(23 - 25k sentences).}
\label{tab:sprlstatistical}
\end{table}
\paragraph{\msprl{}} is a human-curated dataset provided on \sprl{} task. This dataset contains spatial description of real-world images and corresponding \sprl{} annotations~(see Appendix~\ref{sec:sprl}).

\paragraph{\human} did not contain \sprl{} annotations. Hence, we asked two expert volunteers to annotate the story/questions of this dataset. Then another expert annotator checked the annotation and discarded the erroneous ones. As a result, half of this training data is annotated with \sprl{} tags.



\subsubsection{Transfer learning in \sprl{}}

Table~\ref{tab:sprl-role} demonstrates the influence of synthetic supervision in spatial role extraction evaluated on \msprl{} and \human{}. 

We compare the result of SRol model with the previous SOTA, ``R-Inf''~\cite{manzoor2018anaphora}, on \msprl{} dataset. R-Inf uses external multi-modal resources and global inference. 
All of the \bert{}-based SRol models outperform the R-Inf, which shows the power of PLMs for this task. However, since the accuracy of the SRol is already very high, using synthetic supervision shows no improvements compared to the model that only trained with \msprl{} training set for the SRol. In contrast, on \human{}, using synthetic supervision helps the model perform better. Especially,  using \spartun{} increases the performance of the SRol model dramatically, by $15$\%. 

\begin{table}[t]
\centering
\resizebox{\columnwidth}{!}{\begin{tabular}{|llcc|}
\hline
\textbf{Model} & \textit{\textbf{SynSup}}&\textbf{\msprl{}} & \textbf{\spartqa{}-H} \\ \hline
R-Inf & - &80.92& - \\ \hline
SRol & -  &\textbf{88.59}& 55.8 \\ 
SRol & \spartqa{}-A& 88.41 & 57.28\\
SRol & \spartun{}& 88.03 & \textbf{72.43}\\ \hline
\end{tabular}}
\caption{Evaluating spatial role extraction~(SRol) on two \msprl{} and \human{}(\spartqa{}-H) datasets with and without synthetic supervision.}
\label{tab:sprl-role}
\end{table}

In table \ref{tab:sprl-triplet}, we show the result of SRel model (containing spatial relation extraction and spatial relation type classification) for spatial relation extraction, with and without extra supervision from synthetic data. Same as SRol model, extra supervision  from \spartun{} achieves the best result  when tested on \human{}. 

For \msprl{}, we compared the SRel model with R-Inf on spatial relation extraction. As table \ref{tab:sprl-triplet} demonstrates we improve the SOTA by $2.6$\% on F1 measure using \spartun{} as synthetic supervision. Also, model further pretrained on \auto{} gets lower result than model with no extra supervision due to the limited relation expressions used in this data.

\begin{table}[]
\centering
\resizebox{\columnwidth}{!}{\begin{tabular}{|llcc|}
\hline
\textbf{Model} & \textit{\textbf{SynSup}}&\textbf{\msprl{}} & \textbf{\spartqa{}-H} \\ \hline
R-Inf& - & 68.78 & - \\ \hline
SRel & - & 69.12 & S: 48.58\\ 
 &  &  & Q: 49.46\\ 
SRel & \spartqa{}-A & 68.84 & S: 58.32\\ 
 &  &  & Q: 55.17\\ 
SRel & \spartun{} & \textbf{71.23} & S: \textbf{61:53} \\
 &  &  & Q: \textbf{63.22}\\ \hline
\end{tabular}}
\caption{Spatial relation extraction~(SRel) on \msprl{} and \human{}(\spartqa{}-H) with and without synthetic supervision. Since the questions(Q) and stories(S) in \human{} have different annotations~(questions have missing roles), we separately train and test this model on each.}
\label{tab:sprl-triplet}
\end{table}


In conclusion, our experiments show the efficiency of \spartun{} in improving the performance of models on different benchmarks due to the flexible coverage of relation types and expressions.

\section{Conclusion and Future Work}

We created a new synthetic dataset as a source of supervision for transfer learning for spatial question answering (SQA)  and spatial role labeling (\sprl{}) tasks. We show that expanding the coverage of relation types and combinations and spatial language expressions can provide a more robust source of supervision for pretraining and transfer learning. As a result, this data improves the models' performance in many experimental scenarios on both tasks when tested on various evaluation benchmarks. This data includes rules of spatial reasoning and the chain of logical reasoning for answering the questions that can be used for further research in the future.

Moreover, we provide a human-generated dataset on a realistic SQA task that can be used to evaluate the models and methods for  spatial language understanding related tasks in real-world problems. This data is an extension of a previous benchmark on \sprl{} task with spatial semantic annotations. As a result, this dataset contains annotations for both \sprl{} and SQA tasks. 

In future work, we plan to investigate explicit spatial reasoning over text by neuro-symbolic models. Moreover, using our methodology to generate synthetic spatial corpus in other languages or for other types of reasoning, such as temporal reasoning, is an exciting direction for future research.


\section*{Limitations}


Though we aim for a broad coverage of relation types and relations, we collected this from our available resources and spatial lexicons but this is not by any means complete. There can be relations and expressions that are not covered. In particular, the relation expressions are limited to verbs and prepositions. The performance and reasoning ability of our models is improved with transfer learning but this is, certainly, far from the natural language understanding desiderata. Our models are based on large language models and need GPU resources to execute.

\section*{Acknowledgements}
This project is supported by National Science Foundation (NSF) CAREER award 2028626 and partially supported by the Office of Naval Research
(ONR) grant N00014-20-1-2005. Any opinions,
findings, and conclusions or recommendations expressed in this material are those of the authors and
do not necessarily reflect the views of the National
Science Foundation nor the Office of Naval Research. We thank all reviewers for their helpful
comments and suggestions. We also thank Sania Sinha and Timothy Moran for their help in the human data
generation and annotations.



\bibliography{anthology,custom}
\bibliographystyle{acl_natbib}
\clearpage
\newpage

\appendix

\section{Datasets}
\label{sec:appendix}


\subsection{\spartun{}}
\label{sec:appendix-spartun}
As we described in Section~\ref{sec:spartun} to cover more spatial expressions and spatial relation types, we provide an extendable vocabulary of these spatial phenomena. The entire vocabulary of supported relation expressions and entity properties are provided in Figure~\ref{fig:list}. 

\paragraph{Statistic information:}
Each example in \spartun{} contains a story that describe the spatial relation between entities and some questions which ask about indirect relations between entities.
On average, each story contains eight sentences and 91 tokens, which describe ten relations on average. 

We follow \spartqa{} for dataset split. The number of questions in each train, dev, and test sets is provided in Table~\ref{tab:statistical}. YN questions can have two answers "Yes," which is the answer to 54\% of questions, and "No," which is the answer to 46\% of questions. 

FR is a question type with multiple answers.
In below, you can see the percentage of existence of each relation in the whole data: \{ left : 10\%, right:10\%, above: 27\%, below: 26\%, behind: 19\%, front: 10\%, near: 2\%, far: 15\%,  dc: 26\%, ec: 7\%,  po: 0.2\%, tpp: 2\%, ntpp: 10\%, tppi: 3\%, and  ntppi: 8\% \}

\subsection{\resq{}}
\label{sec:sprlqa}
The \resq{} dataset generated over the context of \msprl{} dataset. For each group of sentences ( describing an image), we ask three volunteers (English-speaking undergraduate students) to generate at least four Yes/No questions. On average, they spent 20 minutes generating questions for each group of sentences which, in total, they spent 210 hours generating the whole data. After gathering the data, another undergrad student check the questions and  remove the incorrect ones and keep the rest. The train set is provided on the train set of \msprl{}, and since it does not have a dev set, we split the 32\% of test data (equal to 20\% of the training set) and keep it as the dev set. 50\% of questions in this data are "Yes" and 50\% are "No". The static information of this dataset comes in Table~\ref{tab:statistical}.

To compute the human accuracy we ask two undergraduate students, one from those who create the questions and one new volunteer to answer 100 questions from the test set of \resq{}. In the end a third students grade their answers.

\subsection{\babi{}} 
\label{sec:babi}
\begin{figure}[h]
    \centering
    \includegraphics[width = \linewidth]{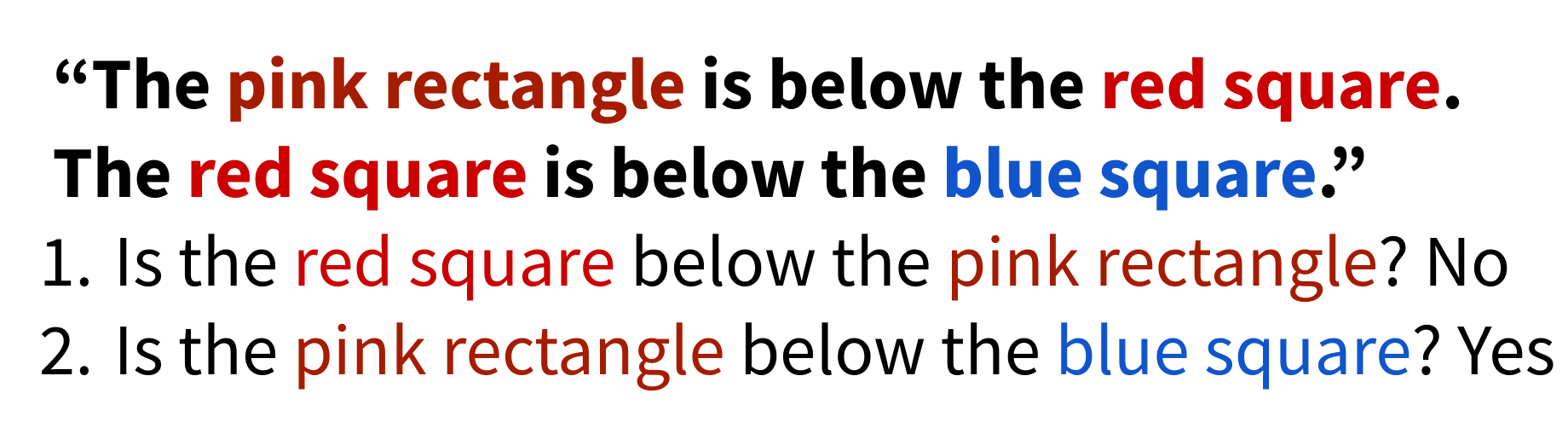}
    \caption{An example of \babi{}}
    \label{fig:babi}
\end{figure}

This dataset is automatically generated data including samples with two sentences describing relationships between three objects and Yes/No questions asking about the existence of a relation between two objects (Fig~\ref{fig:babi}) focuses on multi-hop spatial reasoning question answering. 

 \babi{} task 19, contain questions asking about the directed path from one room to another.
 More statistic information of this dataset comes in table~\ref{tab:statistical}.

\subsection{\spartqa{}}
\label{sec:spartqa}

\begin{figure*}[h]
    \centering
    \includegraphics[width = \linewidth]{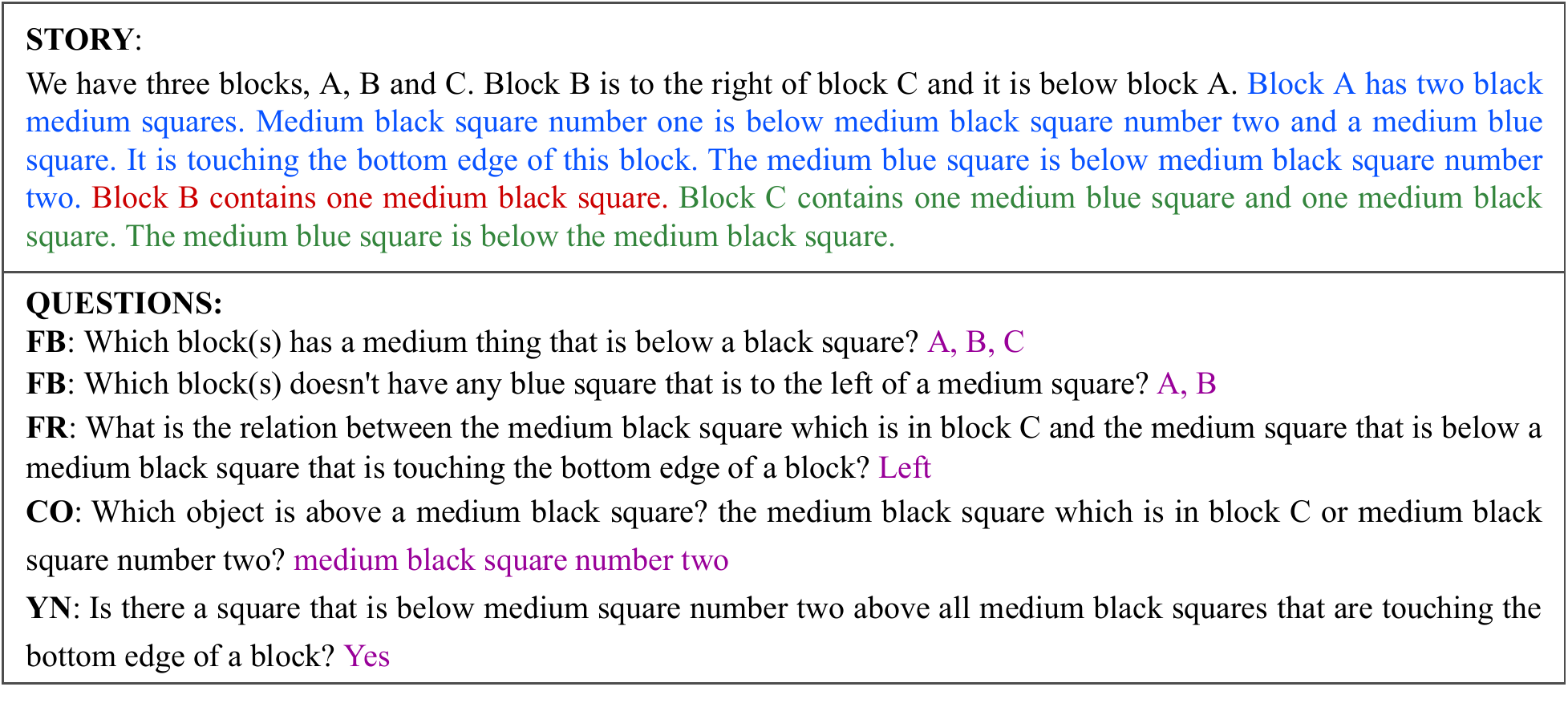}
    \caption{An example of \auto{}}
    \label{fig:spartqa}
\end{figure*}

\paragraph{ \auto} contains more complex textual context (story) and questions requiring complex multi-hop spatial reasoning (e.g. Fig~\ref{fig:spartqa}). This datasets contains one large synthesized~(\auto{}) and a small human-generated~(\human{}) subsets. 

One of the advantages of \spartqa{} is the \sprl{} annotation of whole data (Contexts and Questions) provided with the main dataset. In this work, we also recruited two experts annotator which spent 270 hours annotating  2k sentences in \human{} using WebAnno framework\footnote{\url{https://webanno.github.io/webanno/}}. Then another expert annotator checks their annotation and discards the wrong ones. The statistic information of \spartqa{} comes in Table~\ref{tab:statistical}.

\subsection{\stepgame{}}
\begin{figure}
    \centering
    \includegraphics[width = \columnwidth]{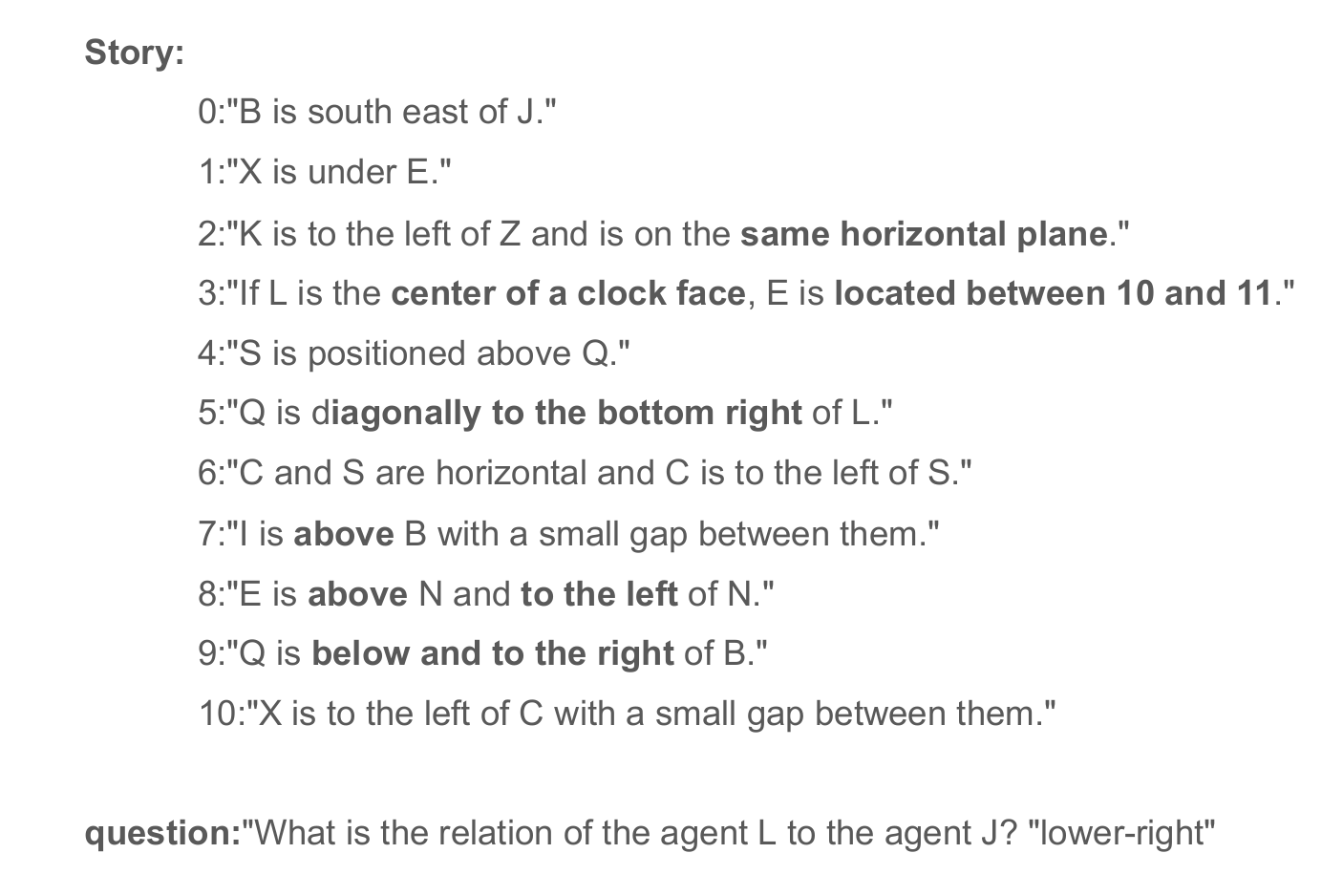}
    \caption{\stepgame{}. An example of questions which need 10 steps of reasoning.}
    \label{fig:stepgame}
\end{figure}

\stepgame{} is another synthesized datasets described in this paper. You can check a sample of this dataset in Figure~\ref{fig:stepgame}. 

\subsection{\msprl{}}
\label{sec:sprl}

\begin{table}[h]
    \centering
    \begin{tabular}{|l|l|l|l|}
        \hline 
        & \text{ Train } & \text{ Test } & \text{ All } \\
        \hline 
        \text{ Sentences } & 600 & 613 & 1213 \\
        \hline 
        \text{ Trajectors } & 716 & 874 & 1590 \\
        \text{ Landmarks } & 612 & 573 & 1185 \\
        \text{ Spatial Indicators } & 666 & 795 & 1461 \\
        \text{ Spatial Triplets } & 761 & 939 & 1700 \\
        \hline
    \end{tabular}
    \caption{\msprl{} statistics.}
    \label{tab:sprl-statistic}
\end{table}

\begin{figure}[h]
    \centering
    \includegraphics[width=\linewidth]{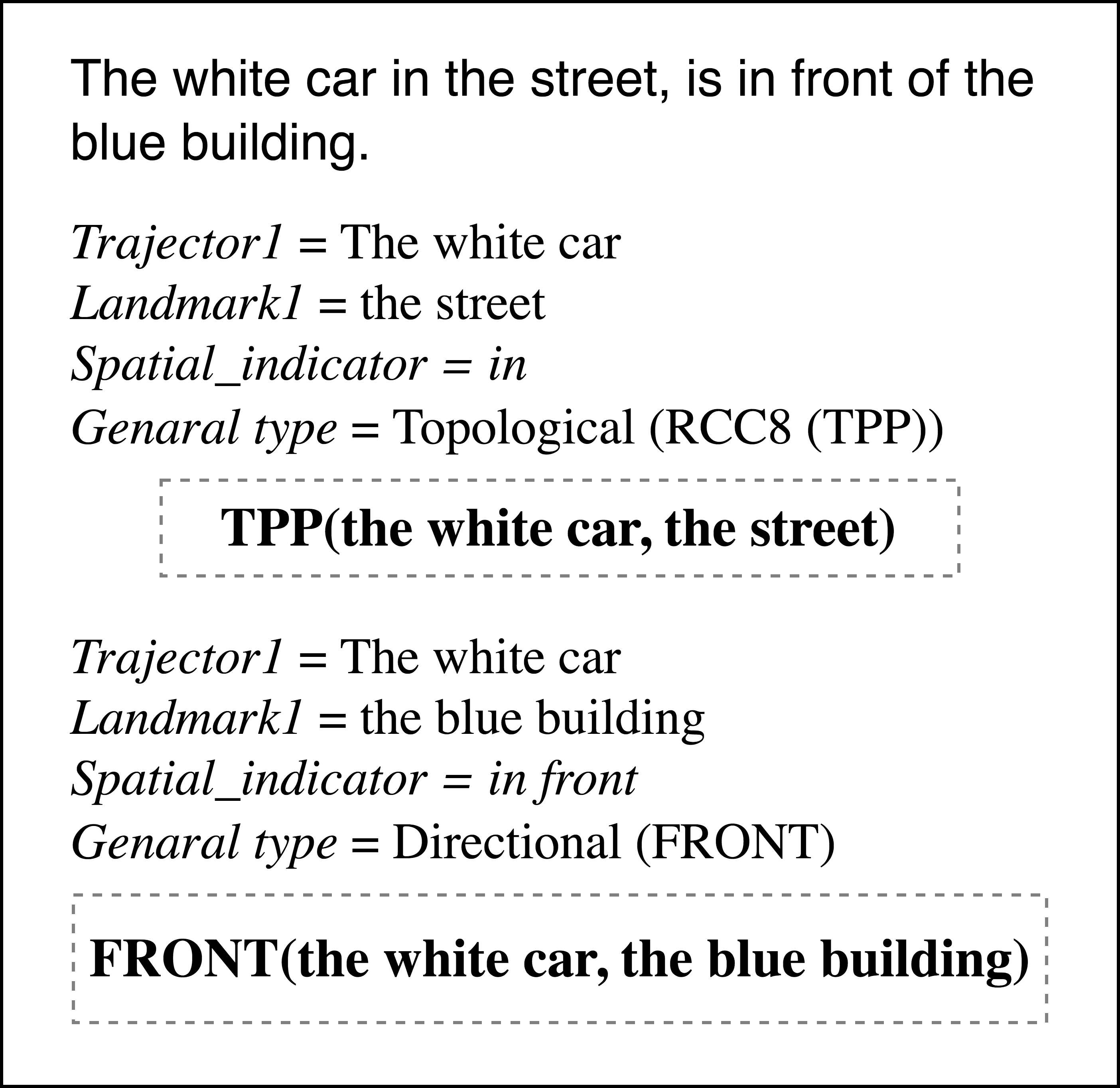}
    \caption{Spatia Role Labeling (SpRL).}
    \label{fig:sprl-ex}
\end{figure}

 \sprl{} is the task of identifying and classifying the spatial arguments of the spatial expressions mentioned in a sentence \cite{kordjamshidi2010spatial}. The \msprl{}\cite{kordjamshidi2017clef} is a dataset provided on \sprl{} task.. The statistic data of this dataset comes in Table~\ref{tab:sprl-statistic}.
A \sprl{} can have following spatial semantic component \cite{zlatev2008holistic} on the static environment, \textbf{trajector} (the main entity), \textbf{landmark}(the reference entity), and \textbf{spatial\_indicator} (the spatial term describing the relationship between trajector and landmark.). The dynamic environment can also have \textit{path}, \textit{region}, \textit{direction}, and \textit{motion}. To understand \msprl{} better you can take a look at Figure~\ref{fig:sprl-ex}. In this figure the spatial value assigned to each spatial triplet can be chosen from Table~\ref{tab:spatial_relations}.

\section{Models and modules}
\label{sec:setup}

We use the huggingFace\footnote{\url{https://huggingface.co/transformers/v2.9.1/model_doc/bert.html}} implementation of pretrained \bert{} base which has 768 hidden dimensions. All models are trained on the training set, evaluated on the dev set, and reported the result on the test set. For training, we train the model until no changes happen on the dev set and then store and use the best model on the dev set. We use AdamW (\cite{loshchilov2017decoupled}) optimizer on all models and modules. 

For SQA tasks we use Focal Loss~\cite{lin2017focal} with $\gamma = 2$. For spatial argument extraction, we use cross-entropy loss for  BIO-tagging, and for spatial relation extraction, we use the summation of loss for each spatial relation and relation type classification part.

\begin{equation}
\begin{split}
    Loss =& \sum\operatorname{CrossEntropyLoss}(\operatorname{p^{\prime}}, \operatorname{y^{\prime}}) 
    \\&+ \operatorname{BCELoss}(\operatorname{p}, \operatorname{y})
\end{split}
\end{equation}

The rest of experimental setting such as number of epochs, batch size, and learning rate are provided in Table ~\ref{tab:experiments-settings}. This settings are chosen after trial and test on the dev set of the target task.

\begin{table}[htbp]
    \centering
    \begin{tabular}{|l|c|c|}
    \hline
         Dataset&YN&FR \\\hline
         \spartun{} &92.83& 93.66\\
         \spartun{}-Simple &90.30& 93.66\\
         \spartun{}-Clock & - & 87.13\\
         \hline
         \spartqa{} & 82.05 & 94.17\\ \hline 
    \end{tabular}
    \caption{Result of \bert{}~(SQA) model trained and test on two synthetic supervision data.}
    \label{tab:sqa-itself}
\end{table}

\begin{table*}[t]
    \small
    \centering
    \resizebox{\textwidth}{!}{
    \begin{tabular}{|ll|cccc|}
        \hline
        \textbf{Experiment}& \textbf{DS}& \textbf{epochs} & \textbf{batch size}& \textbf{learning rate} & \textbf{classifier type} 
         \\ \hline
         PLM & \spartqa{} YN&3&8&8e-06&boolean cls\\
         PLM & \spartqa{} FR&30&8&8e-06&boolean cls\\
         PLM & \stepgame{} &30&4&4e-06&multi-class\\
         PLM & \spartun{} YN&4&8&8e-06&boolean cls\\
         PLM & \spartun{} FR&10&8&8e-06&boolean cls\\\hline{}
        SQA experiments &&&&& \\ \hline 
        \babi{} task 17&&100&4&4e-06& boolean cls\\
        \babi{} task 17&\spartun{}&100&4&4e-06& boolean cls\\
        \babi{} task 17&\spartqa{}&100&4&4e-06& boolean cls\\
        \babi{} task 17&\stepgame{}&100&4&4e-06& boolean cls\\\hline
        \babi{} task 19&&60&4&4e-06& boolean cls\\
        \babi{} task 19&\spartun{}&30&4&4e-06& boolean cls\\
        \babi{} task 19&\spartqa{}&30&4&4e-06& boolean cls\\
        \babi{} task 19&\stepgame{}&30&4&4e-06& boolean cls\\\hline
        \human{} YN&&60&4&4e-06& boolean cls\\
        \human{} YN&\spartun{}&40&4&4e-06& boolean cls\\
        \human{} YN&\spartqa{}&50&4&4e-06& boolean cls\\
        \human{} YN&\stepgame{}&30&4&4e-06& boolean cls\\\hline
        \human{} FR&&40&1&2e-06& boolean cls\\
        \human{} FR&\spartun{}&40&4&4e-06& boolean cls\\
        \human{} FR&\spartqa{}&40&1&2e-06& boolean cls\\
        \human{} FR&\stepgame{}&40&4&4e-06& boolean cls\\\hline
        \stepgame{}&&30&4&4e-0& multi-class\\
        \stepgame{}&\spartun{}&30&4&4e-06& multi-class\\
        \stepgame{}&\spartqa{}&30&4&4e-06& multi-class\\\hline
        \resq{}&&50&4&2e-06& boolean cls\\
        \resq{}&\spartun{}&50&4&4e-06& boolean cls\\
        \resq{}&\spartqa{}& 50&4&4e-06& boolean cls\\
        \resq{}&\stepgame{}&50&4&4e-06& boolean cls\\\hline
        \sprl{} experiments &&&&&\\\hline
        SRol & \spartqa{} &3&1&2e-06&-\\
        SRel & \spartqa{} &5&1&2e-07&-\\
        SRol & \spartun{} &5&1&8e-06&-\\
        SRel & \spartun{} &10&1&2e-07&-\\\hline
        SRol - \human{} &&40&1&2e-05& -\\ 
        SRol - \human{} &\spartqa{}&50&1&2e-06&- \\
        SRol - \human{} &\spartun{}&7&1&4e-07& -\\ \hline
        SRol - \msprl{} &&50&1&2e-06&- \\ 
        SRol - \msprl{} &\spartqa{}&50&1&2e-06&- \\
        SRol - \msprl{} &\spartun{}&50&1&2e-06& -\\ \hline
        SRel - \human{} &&50&1&2e-06&- \\ 
        SRel - \human{} &\spartqa{}&50&1&2e-06&- \\
        SRel - \human{} &\spartun{}&50&1&8e-07& -\\ \hline
        SRel - \msprl{} &&50&1&2e-05&- \\ 
        SRel - \msprl{} &\spartqa{}&70&1&4e-06& -\\
        SRel - \msprl{} &\spartun{}&70&1&6e-06&- \\ \hline

    \end{tabular}
    }
    \caption{The hyperparameters and setups information for each experiment. The first three rows are related to further pretraining model on the synthetic data. These models are used in the other experiments as the further pretrained models.}
    \label{tab:experiments-settings}
\end{table*}

\begin{figure*}[h]
	\centering
	\begin{subfigure}[b]{\linewidth}
		\includegraphics[width=\linewidth]{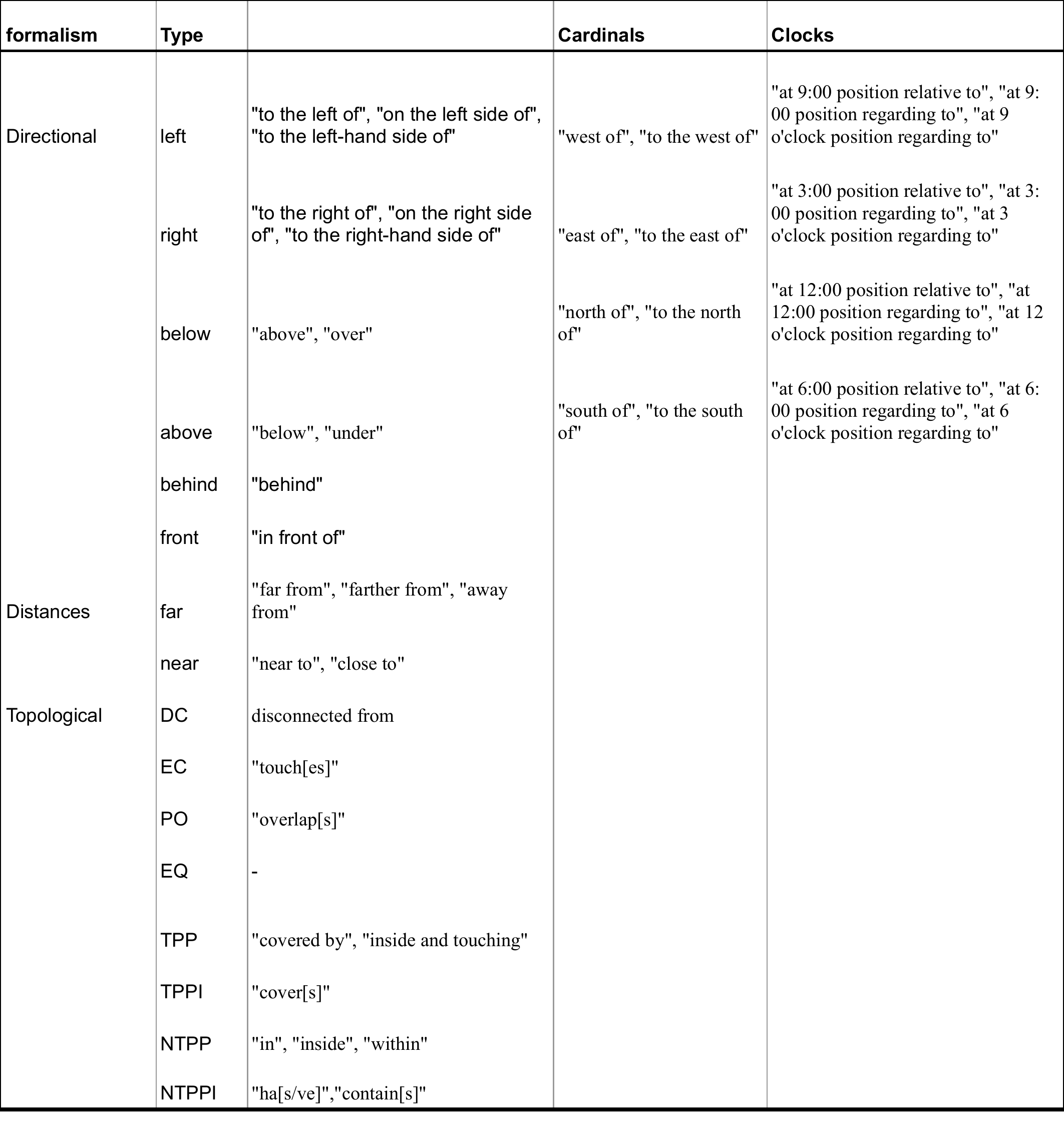}
		\caption{List of relation expression supported in \spartun{}.
		}
		\label{fig:relation_expressions}
	\end{subfigure}
	\begin{subfigure}[b]{0.8\linewidth}
			\includegraphics[width=\linewidth]{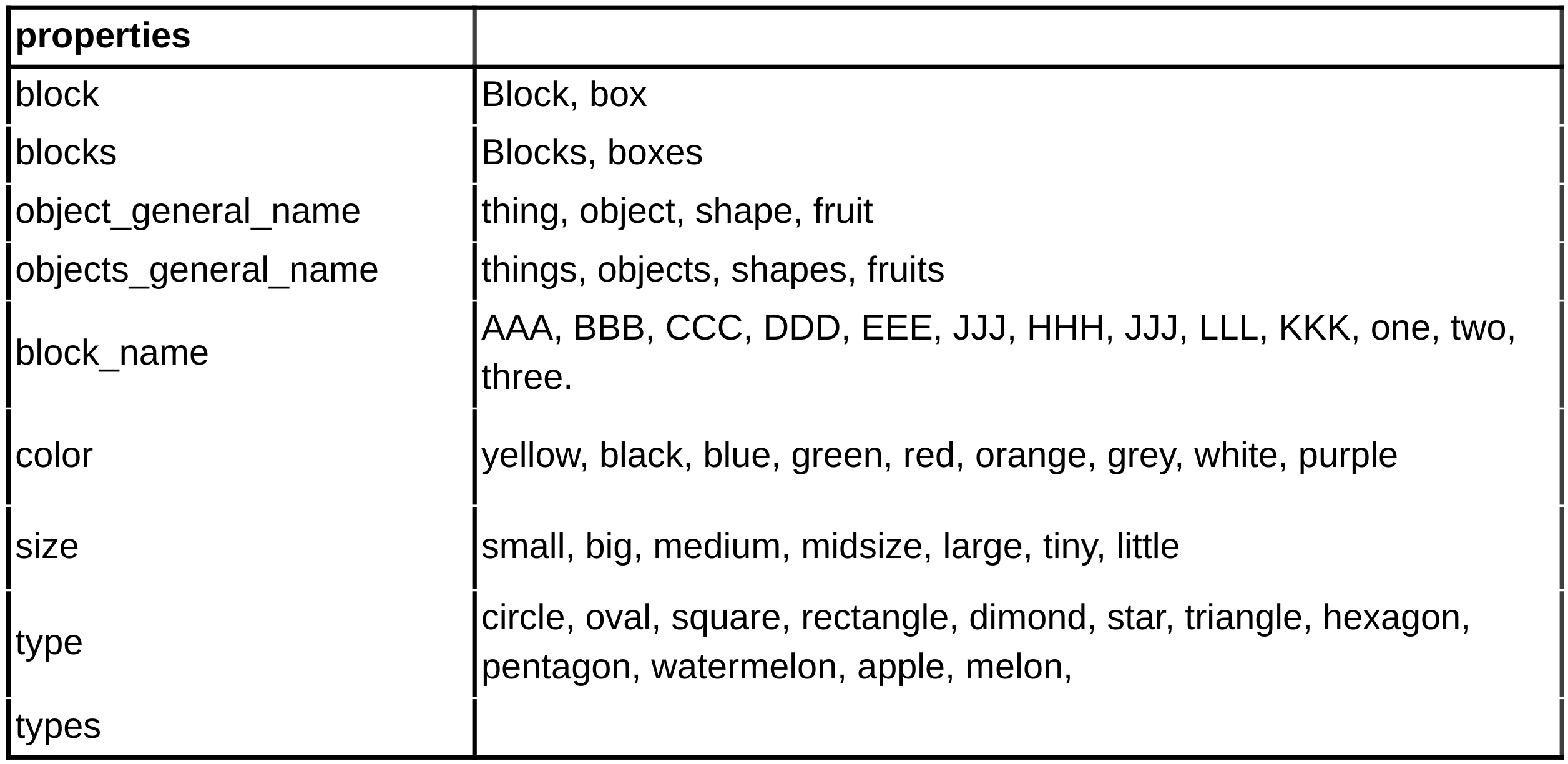}
			\caption{List of entities properties supported in \spartun{}
			}
			\label{fig:obj_properties}
    \end{subfigure}
 	\caption{The supported relation expression and entities properties in \spartun{}, which can easily extended based on the target task.}
	\label{fig:list}
\end{figure*}

Besides, The result of \bert{} model trained on \spartun{} and \spartun{} and tested on the same dataset are provided in Table~\ref{tab:sqa-itself}. \spartun{}-Simple only contains one spatial expresion for each relation types, and \spartun{}-Clock contains all relation expression plus clock expressions~(Column 5 in Table~\ref{fig:relation_expressions}) for relation types.





\subsection{Logic-based spatial reasoner}
\label{sec: prolog}
\begin{figure*}[h]
	\centering
	\begin{subfigure}[b]{\linewidth}
		\includegraphics[width=\linewidth]{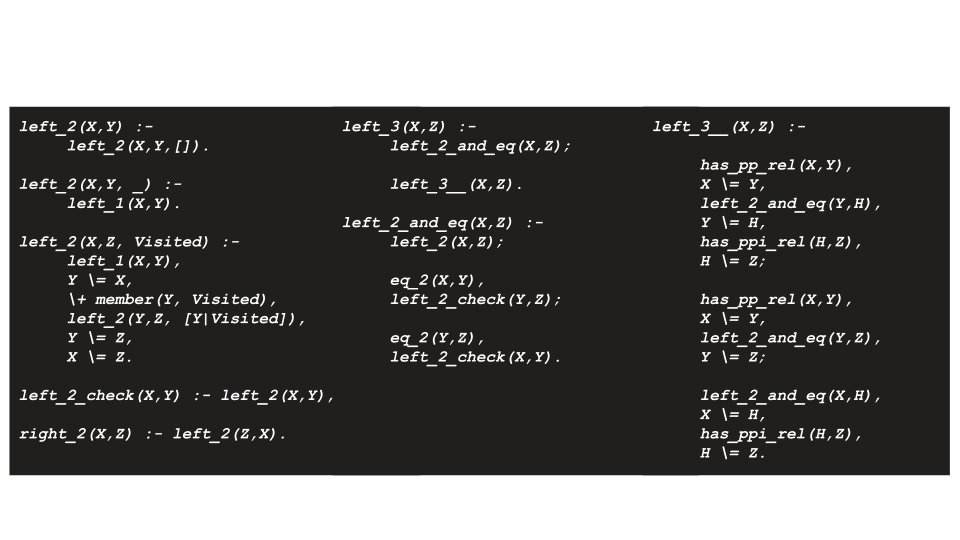}
		\caption{Examaple of implemented rule clauses in Prolog.
		}
		\label{fig:prediction}
	\end{subfigure}
	\begin{subfigure}[b]{\linewidth}
			\includegraphics[width = \linewidth]{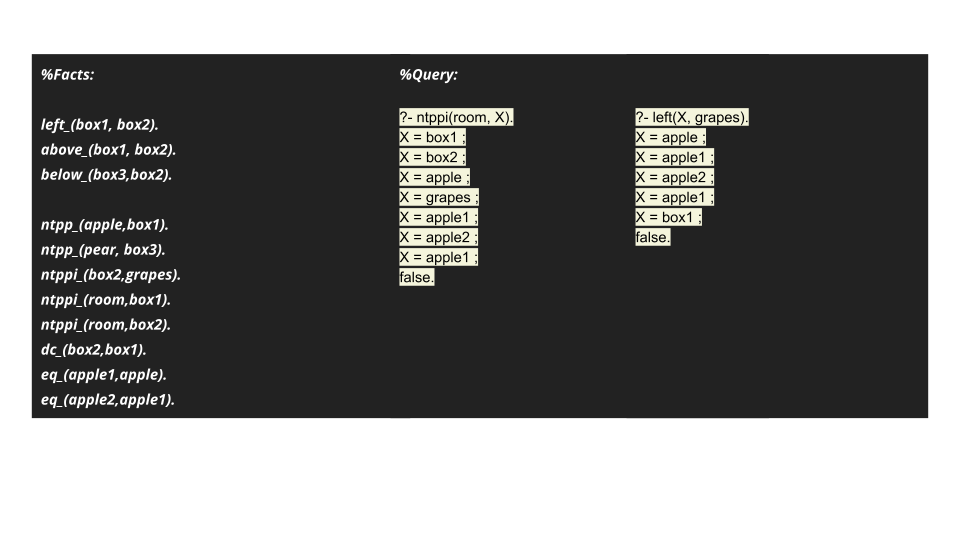}
			\caption{Example of Facts, Query, and answer of implemented model 
			}
			\label{fig:fact}
    \end{subfigure}
 	\caption{Logic-bases spatial reasoner}
	\label{fig:prolog}
\end{figure*}

We consider the logic rules mentioned in Figure~\ref{fig:sp-rules} and in the form of the Horn clauses. we collect the different combinations of spatial relations mentioned in Table~\ref{tab:spatial_relations} and implement the logic-based spatial reasoner. Figure~\ref{fig:prediction} shows an example of some parts of our code on \textit{LEFT} relation. In Figure~\ref{fig:fact},  on the left, some facts are given, and the query ``$ntppi(room,X)$'' ask about all objects that existed in the room. Below each query, there are all possible predictions for them.

\end{document}